\documentclass[runningheads]{llncs}

\usepackage[utf8]{inputenc}
\usepackage[T1]{fontenc}
\usepackage[english]{babel}

\usepackage{graphicx}
\usepackage{hyperref}
\usepackage{xcolor}

\usepackage{acro}
\usepackage{amsmath}
\usepackage{booktabs}
\usepackage{tabularray}
\usepackage{tikz}

\usetikzlibrary{arrows}
\UseTblrLibrary{booktabs}
\usetikzlibrary{positioning}
\usetikzlibrary{external}
\hypersetup{colorlinks=true,allcolors=blue}
\usepackage{cleveref}

\graphicspath{{figs/}}

\DeclareAcronym{AI}{short = AI, long = artificial intelligence}
\DeclareAcronym{BERT}{short = BERT, long = Bidirectional Encoder Representations from Transformers}
\DeclareAcronym{CNN}{short = CNN, long = convolutional neural network}
\DeclareAcronym{DL}{short = DL, long = deep learning}
\DeclareAcronym{GPT}{short = GPT, long = Generative Pretrained Transformer}
\DeclareAcronym{KG}{short = KG, long = knowledge graph}
\DeclareAcronym{KGC}{short = KGC, long = knowledge graph completion}
\DeclareAcronym{LBD}{short = LBD, long = literature-based discovery}
\DeclareAcronym{LINE}{short = LINE, long = Large-scale Information Network Embedding}
\DeclareAcronym{LLM}{short = LLM, long = large language model}
\DeclareAcronym{LM}{short = LM, long = language model}
\DeclareAcronym{MLP}{short = MLP, long = multilayer perceptron}
\DeclareAcronym{NLM}{short = NLM, long = neural language model}
\DeclareAcronym{NLP}{short = NLP, long = natural language processing}
\DeclareAcronym{OBO}{short = OBO, long = Open Biomedical Ontologies}
\DeclareAcronym{PLM}{short = PLM, long = pre-trained language model}
\DeclareAcronym{UMLS}{short = UMLS, long = Unified Medical Language System}
\DeclareAcronym{SLM}{short = SLM, long = statistical language model}

\begin{document}
\title{Recent Advances and Future Directions in Literature-Based Discovery}
%
%
\author{Andrej Kastrin\inst{1}\orcidID{0000-0002-3495-0165} \and
Bojan Cestnik\inst{2,3}\orcidID{0000-0001-8887-5706} \and
Nada Lavrač\inst{2}\orcidID{0000-0002-9995-7093}}
\authorrunning{A. Kastrin et al.}
%
\institute{University of Ljubljana, Ljubljana, Slovenia \\
\email{andrej.kastrin@mf.uni-lj.si} \and
Jo\v{z}ef Stefan Institute, Ljubljana, Slovenia \\
\email{nada.lavrac@ijs.si} \and
Temida d.o.o, Ljubljana, Slovenia \\
\email{bojan.cestnik@temida.si}}
\maketitle
\begin{abstract}
The explosive growth of scientific publications has created an urgent need for automated methods that facilitate knowledge synthesis and hypothesis generation. Literature-based discovery (LBD) addresses this challenge by uncovering previously unknown associations between disparate domains. This article surveys recent methodological advances in LBD, focusing on developments from 2000 to the present. We review progress in three key areas: knowledge graph construction, deep learning approaches, and the integration of pre-trained and large language models (LLMs). While LBD has made notable progress, several fundamental challenges remain unresolved, particularly concerning scalability, reliance on structured data, and the need for extensive manual curation. By examining ongoing advances and outlining promising future directions, this survey underscores the transformative role of LLMs in enhancing LBD and aims to support researchers and practitioners in harnessing these technologies to accelerate scientific innovation.

\end{abstract}

\section{Introduction}

The explosive growth of research publications across various scientific disciplines has led to an overwhelming volume of knowledge, ranging from research articles and monographs to preprints and conference proceedings~\cite{bornmann2021growth}. This proliferation has made it increasingly difficult for researchers to effectively locate, interpret, and synthesize relevant knowledge. As a result, staying current within one's field becomes more challenging, and the risk of missing important findings or inadvertently duplicating existing work rises significantly. Furthermore, the increasing complexity and interdisciplinarity of research further complicate the task of integrating knowledge from multiple sources, and much of the information remains siloed, underutilized, or disconnected. These challenges have led to a rising interest in developing automated methods, particularly those based on \ac{NLP}, to support hypothesis generation and the discovery of novel scientific insights.

A promising approach to address this problem is \ac{LBD}. \ac{LBD}, originally introduced by Swanson~\cite{swanson1986fish} in the mid-1980s, is an approach designed to generate novel research hypotheses by revealing previously overlooked associations between two complementary and non-interactive sets of scientific literatures. It emerged as a response to the growing difficulty of staying abreast of developments across disparate fields and remains a valuable methodology in the face of ever-expanding scholarly output.

The primary motivation of this article is to provide an overview of current methodological challenges in LBD, survey recent scientific advances, and identify future research directions that align LBD with emerging trends in AI and more broadly computational scientific discovery. We limit the scope to the period between 2000 and early 2025, focusing exclusively on state-of-the-art approaches, as earlier methods have already been comprehensively covered in previous surveys~\cite{sebastian2017learning,henry2017literature,thilakaratne2019systematic}.

The article is organized as follows. \Cref{sec:preliminaries} presents the necessary preliminaries and a concise overview of \ac{LBD} research. Recent advances in \ac{LBD} methodologies are examined in \Cref{sec:advances}, followed by a discussion of future research directions in \Cref{sec:future}. The article concludes with a synthesis of key findings in \Cref{sec:conclusion}.

\section{Preliminaries and Background}
\label{sec:preliminaries}

\begin{figure}[ht]
  \centering
  \includegraphics{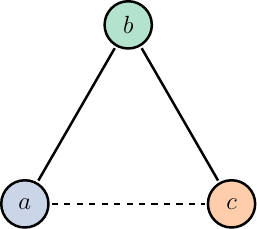}
  \caption{Swanson's ABC model of discovery. When $a$ is related to $b$, and $b$ is related to $c$, it suggests the possibility of an undiscovered indirect relationship between $a$ and $c$.}
  \label{fig:fig-1}
\end{figure}

\ac{LBD} is a subfield of \ac{AI} at the intersection of information retrieval, \ac{NLP} and computational scientific discovery, which is dedicated to automating the scientific discovery process. The early Swanson's approach to \ac{LBD} can be formalized using a generic ABC model (\Cref{fig:fig-1}) that considers two independent literature sets, \( A \) and \( C \)~\cite{swanson1986undiscovered}. In this model, \( a \) represents a source concept, \( c \) is a target concept, and \( b \) serves as a bridge or intermediate concept that connects the two. The key idea is that if \( a \) is associated with \( b \) in one body of literature and \( b \) is associated with \( c \) in another---yet \( a \) and \( c \) have not been directly linked in any publication---there may be a novel, undiscovered relationship between \( a \) and \( c \) worth exploring.

A seminal example of this model is Swanson's~\cite{swanson1986fish} groundbreaking discovery linking dietary fish oil (\( a \)) to Raynaud's disease (\( c \)). He found that Raynaud's disease was associated with reduced blood viscosity in one set of articles, while another set linked high blood viscosity to fish oil. Although no studies at the time had made a direct connection between Raynaud's disease and fish oil, Swanson's hypothesis suggested a new therapeutic use for fish oil, which was later confirmed by clinical research~\cite{digiacomo1989fish}.

In general, \ac{LBD} encompasses two tasks: hypothesis validation and hypothesis generation, which correspond to closed and open discovery modes, respectively. In closed discovery, the process starts with two known elements---a starting concept ($a$) and a target concept ($c$)---and seeks to validate or elaborate the relationship between them by identifying intermediate mechanisms ($b$) that connect $a$ and $c$. Conversely, in open discovery, the process begins with a starting concept ($a$) and aims to uncover previously unrecognized intermediate concepts ($b$) and target concepts ($c$) that could suggest novel research hypotheses.

Despite its pivotal importance, the ABC model exhibits critical limitations that severely constrain the broader applicability of \ac{LBD}. First, scalability remains a pressing challenge. Traditional \ac{LBD} systems were developed for relatively small, curated datasets and are poorly suited to handle the exponential growth of biomedical publications~\cite{thilakaratne2019systematic}. Effectively managing large-scale, heterogeneous corpora demands advanced computational capabilities and methodological innovations that classical \ac{LBD} frameworks were not originally designed to support.

Second, the heavy reliance on structured data sources represents a major constraint. \ac{LBD} approaches have historically depended on controlled vocabularies and ontologies, such as \ac{UMLS}~\cite{bodenreider2004unified}, which facilitate computational access but simultaneously narrow the scope of discovery to well-represented areas~\cite{henry2017literature}. Consequently, current \ac{LBD} systems often exhibit limited flexibility when extracting knowledge directly from unstructured or semi-structured texts, which account for a substantial portion of the scientific literature.

Third, the reliance on extensive manual curation and expert intervention remains a substantial barrier to progress in \ac{LBD}. Traditional LBD workflows necessitate expert involvement at multiple stages, including hypothesis validation, result refinement, and relevance assessment~\cite{sebastian2017learning}. This dependence not only slows the overall discovery process but also poses significant challenges to achieving the scalability and reproducibility required for the broader application of \ac{LBD} tools.

The landscape of LBD is evolving rapidly, but a comprehensive approach to tackling these challenges is essential for realizing its full potential in biomedical research and beyond.

\section{Recent Advances}
\label{sec:advances}

The field of \ac{LBD} has seen notable progress in recent years, driven largely by advancements in machine learning, text mining, and statistical analysis. Research efforts have increasingly harnessed these technologies to develop more effective and sophisticated \ac{LBD} systems. This section reviews three major directions contributing to the recent evolution of \ac{LBD}: \acp{KG}, \ac{DL}, and \acp{LM}.

\subsection{Knowledge Graphs}

\acp{KG} have emerged as a pivotal technology in \ac{NLP}, offering a structured and scalable approach to organizing scientific knowledge. By representing information as networks of entities and their relationships, \acp{KG} enable graph-based reasoning and facilitate the identification of implicit associations across disparate literature sources.

Formally, \acp{KG} are defined as $G = (V, E)$, where $V$ represents the set of vertices (nodes) and $E$ represents edges (links). Relationships in the graph are often modeled as triples $(h, r, t)$, where $h$ (head) and $t$ (tail) are nodes and $r$ is the relation connecting them.

Their construction typically follows two principal methodologies: (i) co-oc\-cur\-rence modeling, where links between entities are established if they co-appear within the same article; and (ii) explicit relation extraction, where semantic relationships are directly identified using specialized \ac{NLP} tools such as SemRep~\cite{rindflesch2003interaction}. Co-occurrence models are widely adopted due to their simplicity and scalability, while relation extraction methods provide greater precision and richer semantic information.\footnote{Co-occurrence refers to the statistical tendency of two terms or concepts to appear together in text (e.g., \emph{fever} and \emph{infection}), without implying a specific semantic or causal relationship. In contrast, a semantic relation denotes a defined and meaningful connection between terms, such as a taxonomic link (e.g., \emph{influenza} as a type of \emph{viral infection}), regardless of how frequently they co-occur.} In particular, co-occurrence-based approaches have gained popularity in \ac{LBD} systems owing to their ease of implementation~\cite{sebastian2017learning,henry2017literature}. Recent approaches also enable the direct construction of \acp{KG} based on predications (subject-relation-object triples) extracted from sources such as PubMed abstracts. Resources such as the \ac{UMLS} and \ac{OBO} offer rich terminological frameworks that enhance the integration and cross-referencing of knowledge.

We approach \ac{LBD} by formulating it as a \ac{KGC} task. \ac{KGC} techniques aim to predict missing information in graphs, either by discovering new edges (link prediction) or by identifying missing nodes (node prediction). Depending on the method used to construct the \ac{KG}, the elements $h$, $r$, and $t$ (as previously defined) may differ: in co-occurrence-based graphs, all three components often represent concepts or terms, whereas in relational databases, $h$ and $t$ denote entities and $r$ represents a predicate, such as \textsc{treats}. Examples include structures like ``Fish oil'' $\rightarrow$ ``Blood viscosity'' $\rightarrow$ ``Raynaud's disease'' for co-occurrence graphs, or ``Fish oil'' $\rightarrow$ \textsc{treats} $\rightarrow$ ``Raynaud's disease'' for relational graphs.

Two main approaches to \ac{KGC} are usually employed: (i) evaluating the plausibility of candidate triples $(h, r, t)$ by assigning a predictive score, and (ii) inferring missing elements by submitting incomplete triples, such as $(h, r)$, $(h, t)$, or $(r, t)$, and predicting the missing component (i.e., predicting $t$ given $(h, r)$, $r$ given $(h, t)$, or $h$ given $(r, t)$).

\subsection{Deep Learning}

In contrast to traditional machine learning methods that rely on features explicitly constructed using domain knowledge, \ac{DL} uses specialized and deep architectures to extract meaningful features from unstructured input, can automatically learn from simple inputs, and extracts task-specific representations of structures.

Crichton et al.~\cite{crichton2020neural} provided compelling evidence that neural network models are highly effective for advancing \ac{LBD}. The authors built upon a \ac{MLP} framework designed for both closed and open discovery tasks, achieving state-of-the-art performance on the PubTator and BioGRID datasets. Their approach begins by generating input representations through node embeddings using the \ac{LINE} algorithm~\cite{tang2015line}, followed by various strategies for combining the embeddings of nodes along a discovery path---structured according to Swanson's ABC model---to construct the input for the neural model.

In closed discovery, the first approach uses a neural model to assign scores to individual $A{-}B$ and $B{-}C$ links, which are then aggregated to evaluate the full $A{-}B{-}C$ path. The second approach combines the embeddings of $A$, $B$, and $C$ into a single input vector, allowing the model to predict a score for the entire path directly, thus removing the need for an explicit aggregation step. In open discovery, the first method similarly scores $A{-}B$ and $B{-}C$ links and aggregates them, but additionally uses an accumulator function to integrate multiple paths leading to the same $C$. The second method employs a \ac{CNN} that processes stacked embeddings of multiple $A{-}B{-}C$ paths, producing a unified score for each $A{-}C$ pair without relying on separate aggregation or accumulation functions. (Unlike conventional \ac{CNN} applications where images are used, the input here is a pseudo-image created by stacking vectorized $A{-}B{-}C$ paths.)

While Crichton et al.~\cite{crichton2020neural} relied on embedding representations for all concepts as model inputs, their method required users to manually construct all possible hypothesis triples prior to evaluation, a process that is both time-consuming and reliant on substantial domain expertise. Addressing this limitation, Cuffy et al.~\cite{cuffy2023exploring} introduced a closed discovery framework that automates the ranking of potential linking $B$ terms for a given $A$ and $C$ pair using a single forward propagation step through the \ac{DL} model. This approach eliminates the need to generate all $A{-}B{-}C$ triples \textit{a priori}, thereby reducing the dependency on domain-specific knowledge and significantly streamlining the \ac{LBD} workflow.

Cuffy et al.~\cite{cuffy2025predicting} introduced a further advancement by reformulating the \ac{LBD} task as the prediction of implicit concept embeddings rather than direct relationship scoring. Instead of classifying triples, their model predicts the embedding of the linking concept ($B$) given the starting ($A$) and target ($C$) concepts. By comparing predicted embeddings against all candidate concepts, the \ac{MLP} model identifies plausible intermediates, demonstrating its effectiveness in systematic knowledge discovery replication.

Beyond general-purpose \ac{LBD} tasks, \ac{DL} has been effectively applied in domain-specific applications, such as drug repurposing. Zhu et al.~\cite{zhu2020extracting} introduced a BioBERT-based model enhanced with entity-aware attention mechanisms for drug-drug interaction extraction, while Gupta et al.~\cite{gupta2021nsga} utilized an NSGA-III-based \ac{CNN} architecture to optimize biomedical search engines. Rather et al.~\cite{rather2017using} further showcased \ac{DL}'s capacity to uncover latent biomedical relationships through word2vec-based embeddings. Taken together, these studies demonstrate the transformative potential of \ac{DL} for \ac{LBD}, facilitating more nuanced knowledge representation and discovery.

\subsection{Language Models}

\Acp{LM} are nowadays regarded as fundamental components of \ac{NLP}, tasked with estimating the probability distributions of linguistic units---such as words, phrases, or sentences---based on their contextual surroundings. The evolution of \acp{LM} can be delineated into several distinct stages: beginning with \acp{SLM}, progressing through \acp{NLM}, advancing to \acp{PLM}, and culminating in the emergence of \acp{LLM}. SLMs utilize basic probabilistic frameworks to model word sequences (e.g., $n$-grams), whereas \acp{NLM} employ neural networks to capture complex syntactic and semantic patterns (e.g., RNNs, LSTMs, transformers). \acp{PLM} leverage large-scale textual corpora and self-supervised learning to encode general linguistic structures and knowledge (e.g., \ac{BERT}, \ac{GPT}).

Here, we focus specifically on how recent developments in \acp{PLM} and \acp{LLM} have been integrated into \ac{LBD} pipelines. While both \acp{PLM} and \acp{LLM} are trained on large corpora using self-supervised methods, \acp{LLM} represent a significant advancement in terms of model size, training data scale, and architectural complexity. Building upon the foundation established by \acp{PLM}, \acp{LLM} offer improved generalization, greater expressivity, enhanced contextual understanding, adaptability, and zero-shot reasoning capabilities---making them particularly well-suited for advanced \ac{LBD} tasks. A comprehensive overview of the historical development of \acp{LM} is provided in recent surveys by Wang et al.~\cite{wang2024history} and Annepaka et al.~\cite{annepaka2025large}.

\acp{PLM} have elevated the quality and scope of \ac{LBD} by integrating deep contextual understanding into \ac{NLP} pipelines. Used both as powerful preprocessing tools and as core components in downstream tasks such as named entity recognition and relation extraction, \acp{PLM} have enabled more accurate and scalable discovery workflows. For example, in our \ac{LBD} approach to drug repurposing for Covid-19~\cite{zhang2021drug}, we employed \ac{BERT}\footnote{Due to its success in general \ac{NLP} tasks, \ac{BERT} has been adapted to various specialized domains, including biomedicine, resulting in models such as BioBERT~\cite{lee2020biobert}, ClinicalBERT~\cite{huang2020clinicalbert}, PubMedBERT~\cite{gu2021domain}, and COVID-Twitter-BERT~\cite{muller2023covid}.} as a preprocessing tool to generate an accurate subset of semantic triples, which were then used to construct a \ac{KG}. \ac{KGC} algorithms were subsequently applied to this graph to predict potential drug repurposing candidates.

Compared to \acp{PLM}, \acp{LLM} exhibit remarkable adaptability, with recent empirical results indicating strong potential for their use as a general-purpose tool to support scientific reasoning~\cite{hope2023computational}. A growing body of evidence reveals a broad range of promising capabilities of \ac{LLM} relevant to the scientific process, including the coherent integration of diverse knowledge concepts, the critical evaluation of existing studies, the generation of scientific hypotheses, and the identification of research gaps within scientific literature~\cite{luo2025llm4sr}. State-of-the-art LLMs, such as BioGPT~\cite{luo2022biogpt} and SciGLM~\cite{zhang2024sciglm} are trained on domain-specific corpora like PubMed and arXiv and are particularly effective at literature retrieval, document summarization, and question answering. They facilitate more efficient access to scientific information by identifying relevant publications, extracting key insights, and synthesizing knowledge across documents.

Specifically, building on the improved reasoning capabilities of \acp{LM}, the \ac{LBD} community has begun developing methods that incorporate richer contextual information to address the limitations of traditional approaches, which are primarily based on the ABC model. Classical \ac{LBD} techniques often fail to capture the nuanced contextual cues considered by human scientists during the ideation process and are largely restricted to predicting pairwise relationships between isolated concepts~\cite{hope2023computational}. To overcome these constraints, Wang et al.~\cite{wang2024scimon} introduced a novel framework, SciMON, which grounds \ac{LBD} in natural language contexts, thereby narrowing the generative space in a more controlled and meaningful way. SciMON optimizes novel research hypotheses by iteratively refining idea suggestions derived from published literature until sufficient novelty is achieved. Unlike traditional models that merely predict conceptual links, SciMON generates complete sentences as outputs, offering a more nuanced and contextually rich representation of scientific knowledge. The authors report that it produces ideas that are both more original and exhibit greater conceptual depth than those generated by GPT-4.

A long-standing limitation of \ac{LBD} has been its restriction to the biomedical domain, primarily due to the widespread availability of the PubMed database and auxiliary knowledge resources (e.g., \ac{UMLS} vocabularies~\cite{bodenreider2004unified}, SemMedDB repository~\cite{kilicoglu2012semmeddb}, PubTator annotations~\cite{wei2024pubtator}), which are freely available and optimized for computational access and analysis. However, \ac{LLM}-powered \ac{LBD} may have a much broader scope of applicability. In particular, Yang et al.~\cite{yang2024large} showed that the majority of published hypotheses in the social sciences can be structured in a manner compatible with the \ac{LBD} framework.

In summary, recent advances in \ac{KG}s, \ac{DL} techniques, and \ac{LM} development have significantly expanded the capabilities of \ac{LBD}. \Cref{tab:tab-1} highlights the principal characteristics of the described approaches. Nevertheless, several key challenges remain, which are discussed in the following section.

\begin{table}[htb]
\centering
\caption{Summary of key strengths and limitations of recent approaches in LBD.}
\begin{tblr}{%
colspec={l|Q[1]|Q[1]},%
}
\toprule
Approach & Strengths & Limitations \\
\midrule
KGs  & Captures complex, heterogeneous associations; enables context-driven subgraph creation. & Requires high-quality semantic annotation; extensive filtering often needed. \\
DL   & Outperforms traditional baselines, especially when input representations are well-optimized. & Interpretability remains a challenge; generalizability may be constrained by data representation. \\
LMs & Offers potential for explainable AI; capable of processing heterogeneous, cross-domain data. & Still emerging; scalability and validation challenges. \\
\bottomrule
\end{tblr}
\label{tab:tab-1}
\end{table}

\section{Future Directions}
\label{sec:future}

Although LBD has made significant advances over the past five years, numerous open challenges remain to be addressed. The following discussion outlines several key areas of ongoing work, reflecting the current focus of our research efforts; however, the list is not intended to be exhaustive.

\subsection{Advancing Interpretability}

Interpretability remains one of the principal challenges associated with the application of \ac{DL} techniques in science~\cite{murdoch2019definitions}. Ensuring interpretability in LBD is not simply an auxiliary feature; it is foundational. While \ac{DL} approaches offer considerable potential for enhancing hypothesis generation from large corpora, their inherent ``black-box'' nature continues to present significant obstacles for scientific domains where transparent reasoning processes are essential. In particular, many \ac{LBD} methods, especially those rooted in Swanson's ABC model, focus primarily on hypothesis generation but often lack mechanisms for explaining the reasoning behind the generated hypotheses. While \ac{DL} systems excel in extracting patterns from literature, they frequently fall short of providing understandable explanations, which symbolic systems have historically offered~\cite{bhuyan2024neuro}.

Traditional strategies, such as employing attention mechanisms or inspecting model coefficients, offer partial solutions by highlighting feature importance or visualizing internal representations~\cite{mersha2024explainable}. However, these approaches often lack a structured reasoning component and thus fall short of delivering full scientific explanatory power. A promising direction is the integration of neuro-symbolic \ac{AI} into \ac{LBD} methodologies. The neuro-symbolic approach aims to combine the pattern recognition capabilities of neural networks with the explicit reasoning structures of symbolic \ac{AI}~\cite{sarker2021neuro,bhuyan2024neuro}. This integration enables models not only to learn from data but also to reason in ways that are inherently interpretable and grounded in logical principles. Neuro-symbolic approaches have already been successfully applied to various \ac{NLP} tasks~\cite{hamilton2024neuro}.

\subsection{Augmenting Data Resources}

One of the principal limitations of current \ac{LBD} applications lies in their restricted use of data resources. Most existing approaches rely primarily on Pub\-Med, often limiting their textual input to article abstracts rather than utilizing the full texts. While abstracts offer a concise summary of findings, they frequently omit critical contextual relationships that could be valuable for complex hypothesis generation, particularly for \acp{LM}. Expanding beyond the biomedical domain to include full-text articles and additional knowledge bases presents a significant opportunity for advancing \ac{LBD}. In particular, new bibliographic databases such as Semantic Scholar~\cite{lo2020s2orc} have emerged as valuable resources. These platforms aggregate extensive metadata, citation networks, and, in some cases, full-text content across a wide range of scientific disciplines, offering richer semantic contexts for discovery processes.

In addition, auxiliary knowledge resources, such as the \ac{UMLS} for the biomedical domain, are of significant importance, particularly during the preprocessing stages of \ac{LBD} (e.g., guiding the extraction of knowledge concepts and the computation of predicates). Although widely integrated into LBD applications for its standardized vocabularies and extensive concept mappings, the use of UMLS is not without limitations. Issues such as term ambiguity and incomplete concept coverage can substantially impact the performance of downstream tasks. For instance, a significant portion of errors in tools like SemRep stem from difficulties in correctly identifying and normalizing biomedical entities using UMLS, accounting for up to 27\% of errors in some evaluations~\cite{kilicoglu2020broad}. FurthermoreIn addition, auxiliary knowledge resources, such as the \ac{UMLS} for the biomedical domain, are of significant importance, particularly during the preprocessing stages of \ac{LBD} (e.g., guiding the extraction of knowledge concepts and the computation of predicates). Although widely integrated into LBD applications for its standardized vocabularies and extensive concept mappings, the use of UMLS is not without limitations. Issues such as term ambiguity and incomplete concept coverage can substantially impact the performance of downstream tasks. For instance, a significant portion of errors in tools like SemRep stem from difficulties in correctly identifying and normalizing biomedical entities using UMLS, accounting for up to 27\% of errors in some evaluations~\cite{kilicoglu2020broad}. Finally, to the best of our knowledge at the time of writing, no comparably well-developed knowledge resource exists outside the life sciences domain. In our experience, the limited adoption of LBD beyond biomedicine is largely due to the greater terminological diversity and, in particular, the absence of standardized ontologies in the humanities and social sciences., to the best of our knowledge at the time of writing, no comparably well-developed knowledge resource exists outside the life sciences domain. In our experience, the limited adoption of LBD beyond biomedicine is largely due to the greater terminological diversity and, in particular, the absence of standardized ontologies in the humanities and social sciences.

\subsection{Refining Benchmarking Practices}

Knuth's~\cite{knuth1984literate} concept of literate programming, which emphasizes that computer programs should be readable and understandable by humans, closely aligns with open science initiatives that stress the importance of standardized practices and tools to ensure research outputs are independently verifiable and can support further scientific progress.

Following the principles of open science, we initiated a project aimed at promoting reproducibility within the field of \ac{LBD}. Existing \ac{LBD} approaches and results often remain difficult to replicate due to the lack of access to original datasets and unresolved programming dependencies. These limitations pose significant barriers to both the theoretical understanding and practical reuse of previously published methods. To address this gap, we have made publicly available benchmark datasets, replicable \ac{LBD} case studies, and a collection of interactive Jupyter Notebooks that transparently document each step of the \ac{LBD} pipeline, including data acquisition, text preprocessing, hypothesis discovery, and evaluation. Furthermore, we provide the \ac{LBD} community with access to standardized benchmark datasets and prototypical \ac{LBD} techniques presented through dockerized Jupyter environments, thereby greatly simplifying replication and extension. All associated materials are openly accessible at \url{https://github.com/akastrin/ida2025lbd}.

\section{Conclusion}
\label{sec:conclusion}

This survey has reviewed the evolution of \ac{LBD} over the past five years. We discussed the growing role of \acp{KG}, advances in \ac{DL} methodologies, and the transformative impact of \acp{PLM} and \acp{LLM} on hypothesis generation and scientific reasoning.

Rapid advances in \ac{AI}, particularly in the development of \acp{LLM}, are reshaping the scientific landscape at an unprecedented pace. These developments open up significant opportunities for treating scientific corpora as dynamic knowledge bases from which novel insights, hypotheses, and ideas can be systematically uncovered. Despite this progress, several fundamental challenges remain unresolved in \ac{LBD}, notably issues related to scalability, dependence on structured data, the need for extensive manual curation, and the limited interpretability of current \ac{DL} approaches.

Recent trends in neuro-symbolic \ac{AI} suggest promising avenues for enhancing both the accuracy and explainability of \ac{LBD} systems. By combining the strengths of \ac{DL} with the reasoning capabilities of symbolic methods, these hybrid approaches aim to deliver more transparent and trustworthy discoveries, thereby enabling broader domain applicability of \ac{LBD} beyond the biomedical sciences.
\begin{credits}
\subsubsection{\ackname} The authors acknowledge the financial support from the Slovenian Research and Innovation Agency through the Knowledge Technologies (Grant No.\ P2-0103), and Methodology for Data Analysis in Medical Sciences (Grant No.\ P3-0154) core research projects, as well as Embeddings-Based Techniques for Media Monitoring Applications (Grant No. L2-50070) research project.

\subsubsection{\discintname}
The authors have no competing interests to declare that are relevant to the content of this article.
\end{credits}

\bibliographystyle{splncs04}
\bibliography{references}
\end{document}